\documentclass{article}

\usepackage{arxiv}

\usepackage[utf8]{inputenc} 
\usepackage[T1]{fontenc}    
\usepackage{hyperref}       
\usepackage{url}            
\usepackage{booktabs}       
\usepackage{amsfonts}       
\usepackage{nicefrac}       
\usepackage{microtype}      
\usepackage{lipsum}
\usepackage{graphicx}
\usepackage{amsmath}

\usepackage{array}
\usepackage{caption}
\graphicspath{ {./images/} }

\usepackage{color,colortbl}
\usepackage{multirow}
\usepackage{xspace}

\title{Evaluation of Segment Anything Model 2: The Role of SAM2 in the Underwater Environment}

\def\mAP{mAP}
\def\AP#1{AP$_{#1}$}

\author{
Shijie Lian\textsuperscript{1,2}, Hua Li\textsuperscript{1}\\
  \textsuperscript{1}Hainan University \\
  \textsuperscript{2}Huazhong University of Science and Technology \\
  \texttt{\{lianshijie, lihua\}}@hainanu.edu.cn \\
}

\begin{document}
\maketitle

\begin{abstract}

With breakthroughs in large-scale modeling, the Segment Anything Model (SAM) and its extensions have been attempted for applications in various underwater visualization tasks in marine sciences, and have had a significant impact on the academic community. 
Recently, Meta has further developed the Segment Anything Model 2 (SAM2), which significantly improves running speed and segmentation accuracy compared to its predecessor.
This report aims to explore the potential of SAM2 in marine science by evaluating it on the underwater instance segmentation benchmark datasets UIIS and USIS10K.
The experiments show that the performance of SAM2 is extremely dependent on the type of user-provided prompts. When using the ground truth bounding box as prompt, SAM2 performed excellently in the underwater instance segmentation domain.
However, when running in automatic mode, SAM2's ability with point prompts to sense and segment underwater instances is significantly degraded.
It is hoped that this paper will inspire researchers to further explore the SAM model family in the underwater domain.
The results and evaluation codes in this paper are available at \url{https://github.com/LiamLian0727/UnderwaterSAM2Eval}.

\end{abstract}


\section{Introduction}

In recent years, large-scale language models (LLMs) such as the Generative Pretraining Transformer (GPT)-4 \cite{achiam2023gpt}, Language Learning for Adaptive Multitasking Architecture (LLaMA) \cite{touvron2023llama}~and Pathways Language Model (PaLM) \cite{JMLR:v24:22-1144}~have sparked a revolution in the field of natural language processing (NLP).
These foundational models exhibit excellent migration capabilities and perform well in numerous open-world language tasks. 
Inspired by the success of LLMs, visual base models such as Contrastive Language-Image Pre-Training model (CLIP) \cite{pmlr-v139-radford21a},  A Large-scale Image and Noisy-text Embedding model (ALIGN) \cite{pmlr-v139-jia21b}, Detection Transformer with Improved Denoising Anchor Boxes v2 (DINOv2) \cite{oquab2023dinov2}, and Segment Anything Model (SAM) \cite{SAM_2023_ICCV}~also emerged. 
The introduction of these foundation models continues to drive researchers' exploration in the field of computer vision.

Among them, the Segment anything model has recently excelled in many segmentation tasks with its excellent encoder-decoder transformation framework and large training dataset SA-1B. 
With fine-tuning or appropriate modifications, it has strong potential in the field of marine science. 
For example, Wang et al. \cite{2024IGRSL..2187712W}~applied SAM to underwater sonar images, through comprehensive and detailed fine-tuning to allow SAM to overcome the challenges of sonar images caused by high noise, low resolution, and complex target shapes.
Zheng et al. \cite{2023arXiv231001946Z}~used SAM to develop an interactive coral labeling tool, producing a large-scale coral video segmentation dataset, CoralVOS.
CoralSCOP \cite{zheng2024coralscop}~also focuses on the coral analysis task, and as the first base model proposed for dense coral segmentation, it shows a powerful generalization ability to unseen coral reef images.
USIS-SAM \cite{lian2024diving}, meanwhile, focuses on more generalized underwater instance segmentation, with excellent segmentation accuracy for common underwater instances (e.g., fish, human divers, underwater robots, etc.).

Following the success of SAM for the image domain, Meta released the SAM2, which is designed to handle image and video segmentation tasks in a unified architecture.
SAM2 surpasses previous capabilities in terms of segmentation accuracy of images with optimal inference speed and strong zero-shot generalization.
In this technical report, we use the underwater instance segmentation task as a case study to analyze the performance of SAM2 in underwater scenarios with the UIIS dataset and the USIS10K dataset.
We observe the following two points: 
\begin{itemize}
  \item 1) When ground truth is used as the prompt for the SAM2, its performance improves significantly compared to SAM and EfficientSAM \cite{xiong2024efficientsam}.
  \item 2) When using SAM2 to automatically generate instance masks, the performance of SAM2 showed significant degradation and is not comparable to state-of-the-art underwater instance segmentation algorithms.
\end{itemize}

\section{Experimental Results}

To validate the performance of SAM, EfficientSAM, and SAM2 in underwater environments, we evaluated them on two benchmark datasets, including underwater instance segmentation dataset, UIIS \cite{Lian_2023_ICCV}, and underwater salient instance segmentation dataset, USIS10K \cite{lian2024diving}.

\textbf{Evaluation Metrics}. We use the standard mask AP metrics \cite{mscoco_2014_ECCV}~as evaluation metrics to fully demonstrate the performance of the model through a series of different IoU thresholds and different scales including \mAP, \AP{50}, \AP{75}, \AP{S}, \AP{M}, and \AP{L}.
In addition, we use the Frames Per Second (FPS) to evaluate model speed. All models were inferred on a single NVIDIA GeForce RTX 4090. When calculating model speed, the image encoder will only run once for each model.

\begin{table*}[ht]
    \begin{center}
    \renewcommand{\arraystretch}{1.2}
    \setlength{\tabcolsep}{2.5mm}
    {\begin{tabular}{c|c|c|c|cccccc}
    \hline\hline
    Method & Prompt & Backbone & FPS & \mAP & \AP{50} & \AP{75} & \AP{S} & \AP{M} & \AP{L}\\
    \hline
    SAM  & 1 Point & ViT-Base & 10.25 & 21.8 & 38.3 & 22.0 & 18.0 & 27.5 & 21.5 \\
    SAM  & 3 Point & ViT-Base &10.19 & 30.6 & 53.9 & 31.1 & 23.6 & 42.5 & 28.2\\
    SAM  & GT Bbox & ViT-Base & 10.37 & 61.7 & 93.8 & 66.6 & 44.1 & 63.0 & 65.8\\
    SAM  & 1 Point & ViT-Huge & 3.84 & 26.5 & 44.5 & 26.9 & 22.3 & 35.0 & 25.6\\
    SAM  & 3 Point & ViT-Huge & 3.84 & 36.0 & 58.0 & 37.6 & 32.0 & 45.4 & 34.7 \\
    SAM  & GT Bbox & ViT-Huge & 3.89 & 65.8 & 96.1 & \textcolor{blue}{78.2} & 47.8 & 64.3 & 73.8 \\
    \hline
    EfficientSAM  & 1 point & ViT-Tiny & \textcolor{red}{27.29} & 19.9 & 37.8 & 18.7 & 20.7 & 28.6 & 18.3\\
    EfficientSAM  & 3 point & ViT-Tiny & 26.81 & 36.0 & 64.1 & 36.4 & 30.3 & 45.7 & 33.6 \\
    EfficientSAM  & GT Bbox & ViT-Tiny & \textcolor{blue}{27.17} & 61.4 & 93.1 & 68.5 & 34.3 & 60.6 & 70.4 \\
    EfficientSAM  & 1 point & ViT-Small & 26.55 & 24.2 & 42.0 & 24.8 & 19.9 & 31.3 & 22.6\\
    EfficientSAM  & 3 point & ViT-Small & 26.13 & 35.5 & 60.5 & 36.5 & 28.9 & 45.0 & 34.0\\
    EfficientSAM  & GT Bbox & ViT-Small & 26.42 & 64.8 & 95.3 & 72.2 & 44.8 & 63.1 & 73.1 \\
    \hline
    SAM2  & 1 Point & Hiera-Tiny & 22.77 & 31.8 & 53.2 & 32.3 & 23.5 & 39.4 & 31.5 \\
    SAM2  & 3 Point & Hiera-Tiny & 22.22 & 40.8 & 68.2 & 41.2 & 28.0 & 47.5 & 43.1 \\
    SAM2 & GT Bbox & Hiera-Tiny & 22.49 & 69.0 & \textcolor{red}{97.3} & \textcolor{blue}{78.2} & \textcolor{red}{54.9} & \textcolor{blue}{68.4} & 76.4\\
    SAM2  & 1 Point & Hiera-Base+ & 19.94 & 33.8 & 55.0 & 35.1 & 26.1 & 41.6 & 34.2\\
    SAM2  & 3 Point & Hiera-Base+ & 19.67 & 43.6 & 71.0 & 46.1 & 29.4 & 50.2 & 45.6 \\
    SAM2 & GT Bbox & Hiera-Base+ & 19.86 & \textcolor{blue}{70.1} & \textcolor{blue}{97.2} & 77.8 & \textcolor{blue}{53.9} & 67.1 & \textcolor{blue}{78.5} \\
    SAM2  & 1 Point & Hiera-Large & 15.24 & 37.8 & 60.2 & 39.9 & 28.6 & 48.1 & 38.1 \\
    SAM2  & 3 Point & Hiera-Large & 15.12 & 48.8 & 76.0 & 51.8 & 33.0 & 54.6 & 52.8 \\
    SAM2 & GT Bbox & Hiera-Large & 15.17 & \textcolor{red}{70.6} & \textcolor{blue}{97.2} & \textcolor{red}{78.7} & 53.8 & \textcolor{red}{69.0} & \textcolor{red}{78.7} \\
    \hline  \hline
    \end{tabular}}
    \end{center}
    \caption{Quantitative comparisons with SAM and EfficientSAM on the UIIS dataset. The \textcolor{red}{red} color is the best and the \textcolor{blue}{blue} color is the second. }
    \label{tab:uiis.comp}
\end{table*}

\subsection{Results of SAM2 on UIIS dataset}

In this subsection, we only evaluate the segmentation capabilities of SAM, EfficientSAM, and SAM2.
In order to exclude other factors as much as possible, we use prompts of the three types, 1 Point, 3 Point, and GT Bbox, to help the model localize the instances to be segmented.
The results can be found in Tab. \ref{tab:uiis.comp}.
Specifically, GT Bbox represents that we use the bounding box in the instance's ground truth as the prompt, 1 Point represents that we use the center of the instance's mask as the prompt, and 3 Point represents that we use the center of the instance's mask and two random boundary points as the prompt.
To minimize the effect of randomness, each experiment was conducted three separate times and the results were averaged.
Since no re-training was required, all tests in the table were done on the test set of the UIIS dataset.

From Tab. \ref{tab:uiis.comp}, it can be observed that SAM, EfficientSAM and SAM2 get the best performance when GT Bbox is used as prompt. Of particular note, SAM2-Hiera-Large achieves a 4.8 AP improvement in mAP while being roughly 5 times faster than SAM-ViT-Huge.
In addition, SAM2-Hiera-Tiny also has a 7.6 AP improvement in mAP compared to EfficientSAM-ViT-Tiny, but is 21\% slower in inference per image.
Therefore, although the inference speed of SAM2 is substantially higher than that of SAM, it can be expected that it still has a lot of space for improvement.

\subsection{Results of SAM2 on USIS10K dataset}

This subsection follows the same experimental setup as subsection 2.1. 
We similarly used three types of prompts, 1 Point, 3 Point, and GT Bbox, to help the model locate the instance to be segmented.
The results are shown in Tab. \ref{tab:usis.comp}.
To minimize the effect of randomness, each experiment was performed three separate times and the results were averaged.
All tests in the table were performed on the test set of the USIS10K dataset.
When using only 1 point as prompt, the SAM2-Hiera-Tiny leads the SAM-ViT-Base in \AP{M} and \AP{S} by 15.9 AP and 16.3 AP, and the EfficientSAM-ViT-Tiny by 12.0 AP and 14.6 AP, demonstrating the SAM2's ability to segment large underwater instance on weak prompts. 

\begin{table*}[ht]
    \begin{center}
    \renewcommand{\arraystretch}{1.2}
    \setlength{\tabcolsep}{2.5mm}
    {\begin{tabular}{c|c|c|c|cccccc}
    \hline\hline
    Method & Prompt & Backbone & FPS & \mAP & \AP{50} & \AP{75} & \AP{S} & \AP{M} & \AP{L}\\
    \hline
    SAM  & 1 Point & ViT-Base & 12.39 & 23.9 & 40.7 & 23.9 & 24.8 & 26.6 & 25.0 \\
    SAM  & 3 Point & ViT-Base & 12.31 & 31.5 & 50.9 & 33.7 & 40.7 & 45.4 & 29.1\\
    SAM  & GT Bbox & ViT-Base & 12.41 & 65.5 & 94.9 & 74.2 & 47.0 & 63.1 & 67.1 \\
    SAM  & 1 Point & ViT-Huge & 4.10 & 29.7 & 47.5 & 30.4 & 27.9 & 30.8 & 30.4\\
    SAM  & 3 Point & ViT-Huge & 4.11 & 38.0 & 57.6 & 40.7 & 39.7 & 48.9 & 36.8 \\
    SAM  & GT Bbox & ViT-Huge & 4.13 & 71.6 & 97.7 & 81.2 & \textcolor{red}{49.9} & 66.2 & 74.2 \\
    \hline
    EfficientSAM  & 1 point & ViT-Tiny & \textcolor{red}{52.53} & 25.6 & 44.2 & 25.5 & 23.5 & 30.5 & 26.7 \\
    EfficientSAM  & 3 point & ViT-Tiny & 51.95 & 42.0 & 66.7 & 44.4 & 42.2 & 50.6 & 40.4 \\
    EfficientSAM  & GT Bbox & ViT-Tiny & \textcolor{blue}{52.41} & 68.3 & 96.0 & 78.7 & 35.9 & 62.8 & 72.1\\
    EfficientSAM  & 1 point & ViT-Small & 48.87 & 26.3 & 44.5 & 27.2 & 22.9 & 29.2 & 27.1 \\
    EfficientSAM  & 3 point & ViT-Small & 48.36 & 35.3 & 55.0 & 38.2 & 44.7 & 48.5 & 33.5\\
    EfficientSAM  & GT Bbox & ViT-Small & 48.95 & 70.3 & 97.2 & 81.2 & 45.4 & 65.4 & 73.5 \\
    \hline
    SAM2  & 1 Point & Hiera-Tiny & 42.87 & 40.3 & 60.7 & 43.1 & 34.4 & 42.5 & 41.3 \\
    SAM2  & 3 Point & Hiera-Tiny & 42.32 & 53.9 & 79.9 & 58.0 & 45.3 & 56.1 & 54.7 \\
    SAM2 & GT Bbox & Hiera-Tiny & 42.76 & 75.6 & \textcolor{red}{98.5} & \textcolor{blue}{88.3} & \textcolor{blue}{49.0} & \textcolor{blue}{69.8} & \textcolor{blue}{78.6}\\
    SAM2  & 1 Point & Hiera-Base+ & 35.12 & 44.2 & 65.2 & 47.7 & 33.2 & 47.3 & 45.8 \\
    SAM2  & 3 Point & Hiera-Base+ & 34.73 & 58.0 & 82.9 & 64.1 & 47.8 & 58.1 & 59.8 \\
    SAM2 & GT Bbox & Hiera-Base+ & 34.65 & \textcolor{blue}{76.7} & \textcolor{blue}{98.2} & \textcolor{red}{89.1} & 50.3 & \textcolor{red}{71.1} &\textcolor{red}{79.8} \\
    SAM2  & 1 Point & Hiera-Large & 22.67 & 47.4 & 70.4 & 50.7 & 40.7 & 48.0 & 48.9 \\
    SAM2  & 3 Point & Hiera-Large & 22.42 & 60.2 & 84.4 & 66.7 & 46.6 & 60.5 & 61.9 \\
    SAM2 & GT Bbox & Hiera-Large & 22.51 & \textcolor{red}{77.2} & 98.0 & 87.5 & 48.5 & 69.2 & 77.6 \\
    \hline  \hline
    \end{tabular}}
    \end{center}
    \caption{Quantitative comparisons with SAM and EfficientSAM on the USIS10K dataset. The \textcolor{red}{red} color is the best and the \textcolor{blue}{blue} color is the second. }
    \label{tab:usis.comp}
\end{table*}

\subsection{Results compared with end-to-end model}

We also compare SAM2 from the automatic model with other end-to-end state-of-the-art methods on the UIIS dataset and the USIS10K dataset.
Specifically, we generate 32$^2$ points uniformly on the image as input prompts to SAM 2, allowing SAM2 to predict multiple masks from them.
Then, low-quality masks with confidence scores below 0.8 are filtered, while duplicate masks are removed by non-maximum suppression. 
In this way, the prediction of the image is obtained.
Finally, to further reduce performance degradation caused by predictions of uninteresting categories, we use Hungarian matching between the SAM2 predicted masks and the ground truth masks, and then calculate the AP only for the successfully matched predicted masks. 

As can be seen in Tab. \ref{tab:sota.comp}, at this time, the performance of SAM2 shows a significant degradation. 
In addition, the inference speed of SAM2 decreases substantially due to the input more than 900 point prompts on each image. Specifically, on the USIS10K dataset, the FPS of SAM2-Hiera-Tiny, SAM2-Hiera-Base+, and SAM2-Hiera-Large are 1.53, 1.41, and 1.32, respectively, whereas on mAP they lag behind SAM-ViT-Huge by 26.1 AP, 22.6 AP, and 24.9 AP.

\begin{table*}[t]
    \begin{center}
    \renewcommand{\arraystretch}{1.2}
    \setlength{\tabcolsep}{3mm}
    {\begin{tabular}{c|c|ccc|ccc}
    \hline\hline
    \multirow{2}{*}{Method}  & \multirow{2}{*}{Backbone}  &\multicolumn{3}{c|}{UIIS Dataset} & \multicolumn{3}{c}{USIS10K Dataset}\\ \cline{3-8}
     &  &  \mAP & \AP{50} & \AP{75} & \mAP & \AP{50} & \AP{75}\\
    \hline
    Mask RCNN \cite{MASK_RCNN_2017} & ResNet-101  & 23.4 & 40.9 & 25.3 & 32.4 & 49.6 & 35.4\\
    RDPNet \cite{RDPnet_2021_TIP} & ResNet-101  & 20.6 & 38.7 & 19.4 & 39.3 & 55.9 & 45.4\\
    WaterMask \cite{Lian_2023_ICCV} & ResNet-50  & 26.4 &43.6 &28.8 & 37.7 & 54.0 & 42.5 \\
    WaterMask \cite{Lian_2023_ICCV}  & ResNet-101  & 27.2 & 43.7 &29.3 & 38.7 & 54.9 & 43.2\\
    USIS-SAM \cite{lian2024diving}  & ViT-H  & \textcolor{blue}{29.4} &\textcolor{blue}{45.0} & \textcolor{blue}{32.3} & \textcolor{blue}{43.1} & \textcolor{blue}{59.0} & \textcolor{blue}{48.5}\\
    \hline
    SAM (automatic)*  & ViT-Base  & 14.8 & 23.4 & 16.3  & 26.5 & 37.5 & 29.9\\
    SAM (automatic)*  & ViT-Huge  & \textcolor{red}{40.3} & \textcolor{red}{61.4} & \textcolor{red}{43.9}  & \textcolor{red}{52.9} & \textcolor{red}{72.4} & \textcolor{red}{60.8}\\
    SAM2 (automatic)* & Hiera-Tiny  & 7.2 & 9.9 & 8.2  & 18.0 & 21.7 & 20.0\\
    SAM2 (automatic)* & Hiera-Base+  & 15.7 & 20.6 & 18.0  & 30.3 & 36.5 & 34.1\\
    SAM2 (automatic)* & Hiera-Large & 17.9 & 23.8 & 19.7  & 28.0 & 33.1 & 31.1\\
    \hline  \hline
    \end{tabular}}
    \end{center}
    \caption{Quantitative comparisons with state-of-the-arts on the UIIS datasets and USIS10K dataset. The \textcolor{red}{red} color is the best and the \textcolor{blue}{blue} color is the second. * indicates that we will use the Hungarian matching algorithm between the predicted mask and the ground truth mask, and then only calculate the APs of the predicted masks that are successfully matched.}
    \label{tab:sota.comp}
    \vspace{-5mm}
\end{table*}

\subsection{Visualization results}

\begin{figure}[!t]
  \centering
  \includegraphics[width=1\linewidth]{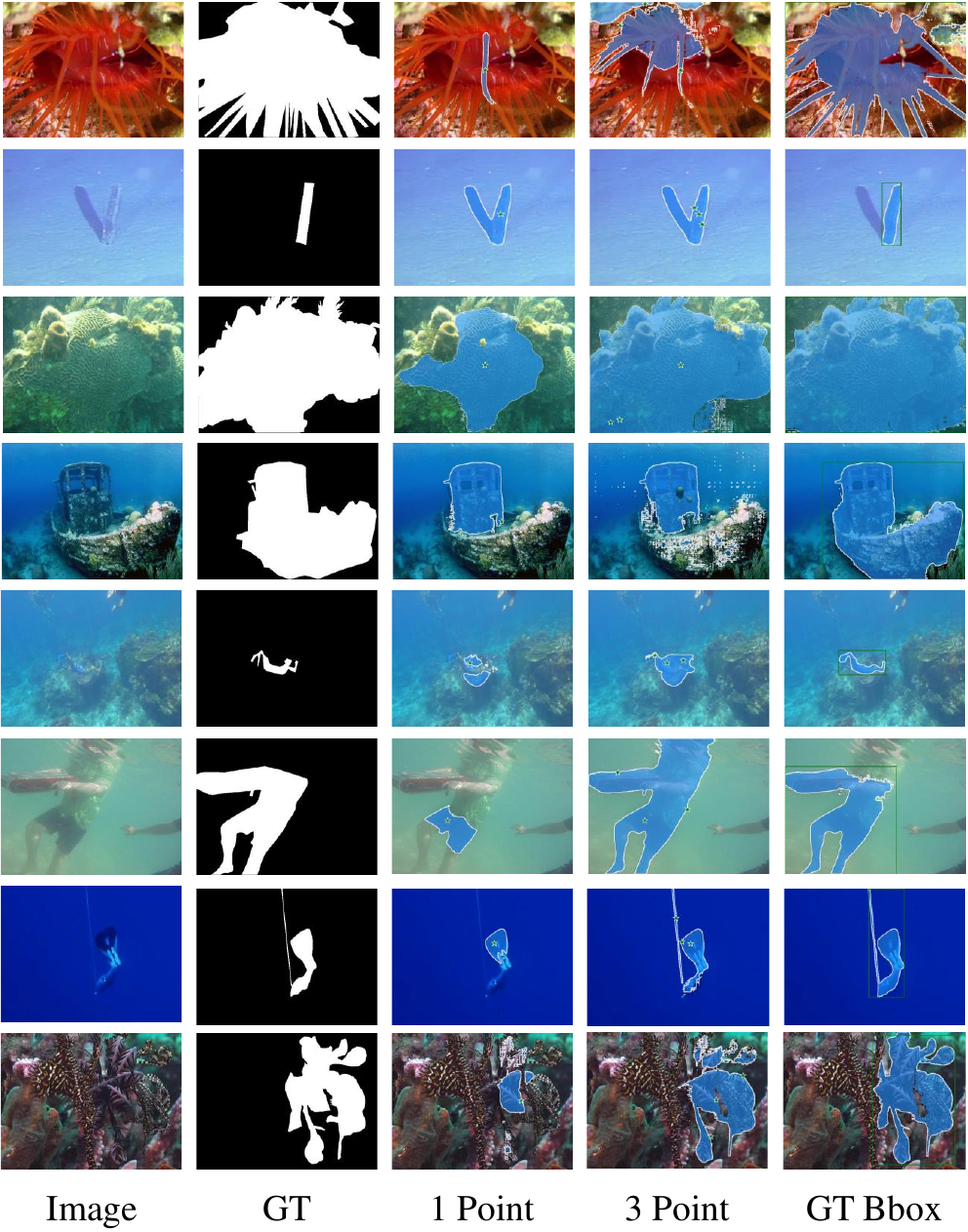}
  \vspace{-5mm}
  \caption{ SAM2 Visualisation results at different prompts.}
  \label{fig:show}
\end{figure}

We also present some visualization results for SAM2 in Fig \ref{fig:show} to show the performance of SAM2 at different prompts.
It can be seen that when dealing with underwater instances with visual ambiguity phenomena (e.g., rows 2 and 6), SAM2 tends to segment out the wrong masks. When processing large objects, SAM2 sometimes generates a large amount of noise at the edges (e.g., rows 3 and 4).

\section{Conclusion}\label{sec:dis}

In this work, we conduct a preliminary investigation of the performance for SAM2 in the field of underwater segmentation. Based on experiments on the UIIS dataset and the USIS10K dataset, we observe:
\begin{enumerate}
    \item The performance of SAM2 is largely dependent on the type and quality of the input prompts, and when the type of prompts is constant, the difference in performance between different backbones of SAM2 is not significant. 
    \item When automated inference without user-specified prompt, the performance of SAM2 shows a significant degradation. Therefore, how to design a reliable object detection module as a prompt generator for SAM2 will be the focus of future research in this area.
\end{enumerate}
In addition, due to the scarcity of underwater video segmentation datasets, this report doesn't evaluate the performance of SAM2 on the underwater video instance segmentation task. However, based on SAM2's excellent performance on underwater 2D instance segmentation, we speculate that SAM2 can be a powerful annotation tool for underwater video instance segmentation dataset and help the development of this field.

\clearpage
\bibliographystyle{splncs04}
\bibliography{references}
\end{document}